\documentclass{SCIS2025}
\usepackage{amsmath, amssymb, amsfonts}

\usepackage{algorithm}
\usepackage{algorithmic}

\usepackage{graphicx}
\usepackage{caption}
\usepackage{subcaption}
\usepackage{booktabs}
\usepackage{multirow}
\usepackage{array}
\usepackage{makecell}
\usepackage{color, xcolor}
\usepackage[percent]{overpic}

\usepackage{textcomp}
\usepackage{epstopdf}
\usepackage{url}
\usepackage{colortbl}
\usepackage{verbatim}
\usepackage{cite}
\usepackage{bbding}
\usepackage{mdframed}
\usepackage{arydshln}
\usepackage{soul}
\usepackage[thicklines]{cancel}
\usepackage{pifont}

\usepackage[capitalize]{cleveref}
\crefname{section}{Sec.}{Secs.}
\Crefname{section}{Section}{Sections}
\Crefname{table}{Table}{Tables}
\crefname{table}{Tab.}{Tabs.}
\Crefname{equation}{Equation}{Equations}
\crefname{equation}{Eqn.}{Eqns.}

\makeatletter
\DeclareRobustCommand\onedot{\futurelet\@let@token\@onedot}
\def\@onedot{\ifx\@let@token.\else.\null\fi\xspace}

\def\eg{\emph{e.g}.} 
\def\ie{\emph{i.e}.} 
 
\def\etc{\emph{etc}.} \def\vs{\emph{vs}.}

\makeatother

\def \robotlogo {\raisebox{-0.1\height}{\includegraphics[height=0.95\baselineskip]{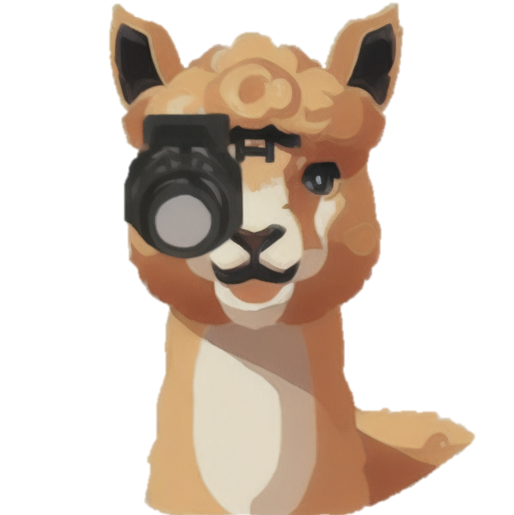}}}

\newcommand{\stdvu}[1]{\scriptsize{\color{darkgray}(#1)}}

\graphicspath{{figures/}}

\begin{document}

\ArticleType{RESEARCH PAPER}
\Year{2025}
\Month{}
\Vol{}
\No{}
\DOI{}
\ArtNo{}
\ReceiveDate{}
\ReviseDate{}
\AcceptDate{}
\OnlineDate{}
\AuthorMark{}
\AuthorCitation{}

\title{\robotlogo{} Myriad: A Large Multimodal Model Applying Vision Experts for Industrial Anomaly Detection}{Myriad: A Large Multimodal Model Applying Vision Experts for Industrial Anomaly Detection}

\author[1,2$^\dagger$]{Yuanze LI}{}
\author[1,2$^\dagger$]{Haolin WANG}{}
\author[1,2]{Shihao YUAN}{}
\author[1,2]{Ming LIU}{{csmliu@outlook.com}}
\author[1]{\\Debin ZHAO}{}
\author[3]{Yiwen GUO}{}
\author[2]{Chen XU}{}
\author[2]{Guangming SHI}{}
\author[1,2]{Wangmeng ZUO}{}

\address[1]{Faculty of Computing, Harbin Institute of Technology, Harbin, 15001 China}
\address[2]{Pazhou Lab Huangpu, Guangzhou, 510555 China}
\address[3]{Independent researcher}

\abstract{
Due to the training configuration, traditional industrial anomaly detection (IAD) methods have to train a specific model for each deployment scenario, which is insufficient to meet the requirements of modern design and manufacturing.
On the contrary, large multimodal models~(LMMs) have shown eminent generalization ability on various vision tasks, and their perception and comprehension capabilities imply the potential of applying LMMs on IAD tasks.
However, we observe that even though the LMMs have abundant knowledge about industrial anomaly detection in the textual domain, the LMMs are unable to leverage the knowledge due to the modality gap between textual and visual domains.
To stimulate the relevant knowledge in LMMs and adapt the LMMs towards anomaly detection tasks, we introduce existing IAD methods as vision experts and present a novel large \underline{m}ultimodal model appl\underline{y}ing vision expe\underline{r}ts for \underline{i}ndustrial \underline{a}nomaly \underline{d}etection~(abbreviated to {Myriad}).
Specifically, we utilize the anomaly map generated by the vision experts as guidance for LMMs, such that the vision model is guided to pay more attention to anomalous regions.
In this way, the training of LMMs is associated with the understanding of anomalies, making the training smoother.
It is also worth noting that the LMMs can be naturally extended to various settings (\eg, zero-shot, few-shot, and \etc) by applying different vision experts without architecture modification.
Then, the visual features are modulated via an adapter to fit the anomaly detection tasks, which are fed into the language model together with the vision expert guidance and human instructions to generate the final outputs.
Extensive experiments on MVTec-AD, VisA, and PCB Bank benchmarks demonstrate that our proposed method not only performs favorably against state-of-the-art methods under one-class and few-shot settings, but also inherits the flexibility and instruction-following ability of LMMs in the field of IAD.
Source code and pre-trained models are publicly available at \url{https://github.com/tzjtatata/Myriad}.
}


\maketitle

\begin{figure*}[t]
    \centering
    \small
    \begin{tabular}{cc}
        \begin{tabular}{c}
            \includegraphics[width=0.4\linewidth]{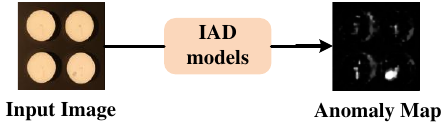} 
            \label{fig:1a}\\
            (a) The framework of existing IAD methods \\
            \includegraphics[width=0.4\linewidth]{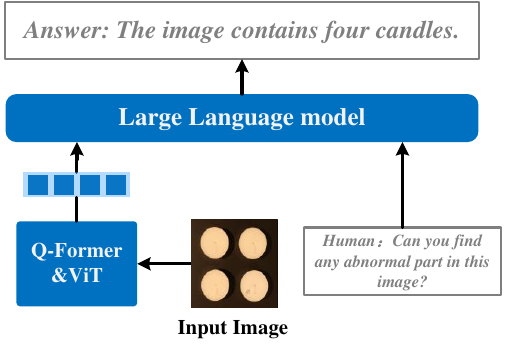} \\
            (b) The framework of MiniGPT-4
        \end{tabular}
        \begin{tabular}{c}
            \includegraphics[width=0.54\linewidth]{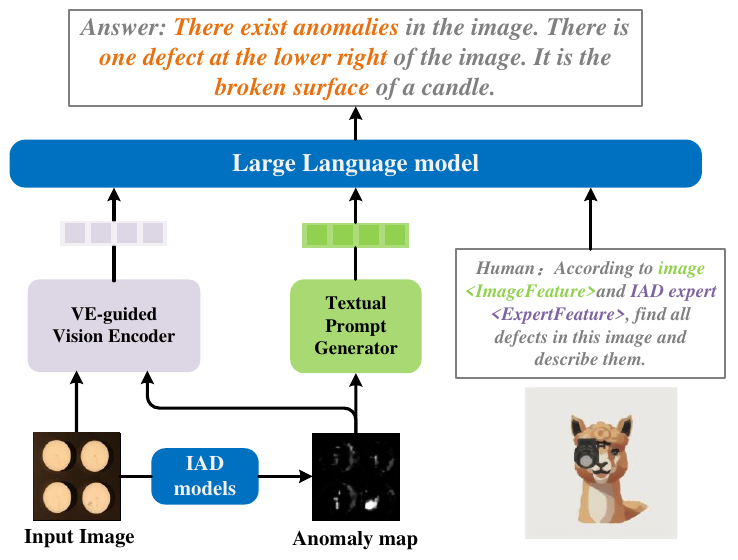} \\
            (c) The framework of our proposed Myriad
        \end{tabular}
    \end{tabular}
    \caption{Existing IAD methods are limited in predicting anomaly maps and anomaly scores without comprehension descriptions (a) while LMMs like MiniGPT-4 cannot well-generate IAD-related descriptions, (b) By incorporating pre-trained IAD models as vision experts, our Myriad can perceive IAD domain knowledge via the introduced vision expert-guided vision encoder (\cref{sec:veguided}) and vision expert guidance adapter(\cref{sec:veguidance}). Our Myriad provides not only favorable anomaly detection accuracy but also instruction-following capability.}
    \label{fig:overview}
\end{figure*}

\section{Introduction}

Industrial anomaly detection~(IAD) focuses on identifying and localizing manufacturing defects, playing a critical role in ensuring reliable, uninterrupted industrial processes. By automatically detecting anomalies, IAD enables timely intervention, efficient maintenance, and process optimization, ultimately boosting productivity and minimizing downtime. In recent years, substantial research attention has been devoted to IAD, resulting in the release of numerous datasets~\cite{mvtec,visa, ds_BTAD, ds_MPDD, ds_MIAD} and methods~\cite{simplenet,PaDiM,patchcore,aprilgan,DeSTSeg,DiffAD,diffusionAD,lafite}.

Despite these efforts, conventional IAD approaches still suffer from a critical limitation: they typically require a separate model for each deployment scenario. Once trained, the model’s input/output format and overall capabilities are fixed, making it difficult to adapt to diverse industrial requirements or to generate outputs based on varying instructions. Consequently, such methods struggle to meet the dynamic demands of modern design and manufacturing.

To overcome these limitations, large multimodal models (LMMs) have emerged as a promising alternative. 
Building on top of powerful large language models (LLMs), such as LLaMA~\cite{llama} and Vicuna~\cite{vicuna}, LMMs inherit strong comprehension skills and the ability to follow instructions. By coupling these LLMs with vision models and vision-to-language conversion modules, LMMs gain an advanced understanding of visual inputs, enabling them to excel in a variety of tasks—ranging from image captioning to visual grounding. 
With robust perception and comprehension abilities, LMMs offer a promising pathway to overcome the shortcomings of conventional IAD. By leveraging LMMs, it may be possible to develop next-generation IAD systems that can adapt flexibly to new instructions and diverse deployment environments, meeting the demands of modern design and manufacturing.

Large multimodal models (LMMs) possess a wealth of knowledge about industrial anomaly detection (IAD) in the textual domain (please refer to the left part of \cref{fig:vicuna-candle}). However, they struggle to leverage this knowledge for visual data, primarily due to the inherent gap between textual and visual modalities.
Specifically, applying LMMs to IAD tasks directly is infeasible from two aspects, \ie,
i)~existing LMMs are typically trained for general purposes, leaving their anomaly-oriented knowledge unlinked to the vision modality (please refer to \cref{sec:discussion} for a more detailed illustration), and
ii)~the limited amount of pairwise vision-language data for IAD makes a full fine-tuning of LMMs impractical.
To stimulate the knowledge in LMMs and adapt the LMMs to anomaly detection tasks, we propose regarding existing IAD models as vision experts (VE) of anomaly detection tasks, and establish a novel large \textbf{m}ultimodal model appl\textbf{y}ing vision expe\textbf{r}ts for \textbf{i}ndustrial \textbf{a}nomaly \textbf{d}etection (abbreviated to \textbf{Myriad}).
By combining the specialized capabilities of these VEs with the extensive knowledge embedded in LMMs, Myriad bridges the modality gap and effectively adapts LMMs for industrial anomaly detection.

To begin with, following MiniGPT-4~\cite{minigpt4}, we stack the pre-trained ViT backbone from EVA-CLIP~\cite{evaclip} and the Q-Former from BLIP-2~\cite{blip2} to form a fundamental vision module.
On this basis, we construct a vision-expert (VE)-guided vision encoder that produces \textit{IAD-suitable visual features} under the the anomaly maps generated by vision experts. 
To adapt these features for IAD, we deploy a Low-Rank Residual Adapter~(LoRRA) between the ViT backbone and the Q-Former.
For leveraging the auxiliary visual information from the anomaly maps, we introduce a trainable visual prompt generator that derives a set of IAD expert query tokens. 
These tokens extend the Q-Former’s pre-trained query tokens, which remain frozen (along with the ViT and Q-Former parameters) to avoid overfitting the limited training data.
By incorporating the anomaly maps, the Q-Former is guided to pay more attention to regions with higher anomaly scores. This strategy compensates for the lack of IAD-specific knowledge in pre-trained language–vision models (LMMs), ultimately enhancing anomaly detection performance.

Beyond the VE-guided visual features, we also provide two extra inputs to the language model. 
First, we feed the anomaly map itself into the model through a textual prompt generator, offering additional cues such as positional information that purely appearance-based visual features might overlook. 
Second, we design a linguistic instruction that mirrors the logic and workflow of anomaly detection. 
Both the visual features and vision expert prompts are embedded into this instruction via placeholders~(shown in \cref{fig:overview}), forming the final input to the LLM. 
By integrating the anomaly maps, visual features, and carefully crafted instructions, our pipeline not only boosts anomaly detection capabilities but also supports various settings (e.g., zero-shot, few-shot, and \etc) and paves the way for interactive, instruction-driven analysis.

To evaluate Myriad, we conduct extensive evaluations on the MVTec-AD~\cite{mvtec},  VisA~\cite{visa}, and PCB Bank~\cite{pcb_bank} datasets. The experimental results show that Myriad not only outperforms both domain-specific vision experts and state-of-the-art industrial anomaly detection methods, but also inherits the flexibility and instruction-following abilities of large multimodal models (LMMs) in the context of IAD. Qualitative results reveal that Myriad can follow human instructions to generate detailed descriptions of defects and discuss potential causes.

Our contributions can be summarized as follows.
\begin{itemize}
    \item A new framework, Myriad, is proposed for industrial anomaly detection (IAD). By introducing existing IAD models as vision experts into LMMs, Myriad effectively stimulates the IAD-relevant textual domain knowledge in LMMs and is adapted to the anomaly detection task. Besides, Myriad inherits the flexibility and instruction-following ability of LMMs and can produce comprehensive descriptions of anomalies following diverse human instructions.
    \item For extracting proper visual features and bridging domain gaps between general-purpose LMMs and anomaly detection tasks, we design a vision expert-guided vision encoder. In particular, a visual query generated from the vision expert guidance is embedded into the Q-Former, and a Low-Rank Residual Adapter~(LoRRA) is also deployed to coordinate with the visual feature modulation process.
    \item Extensive experiments show that our proposed Myriad can appropriately receive prior knowledge from vision experts and outperforms state-of-the-art methods on MVTec-AD~\cite{mvtec}, VisA~\cite{visa}, and PCB Bank~\cite{pcb_bank} datasets.
\end{itemize}

\begin{figure*}[t]
    \small
    \centering
    \renewcommand\tabcolsep{2pt}

    \includegraphics[width=1.0\linewidth]{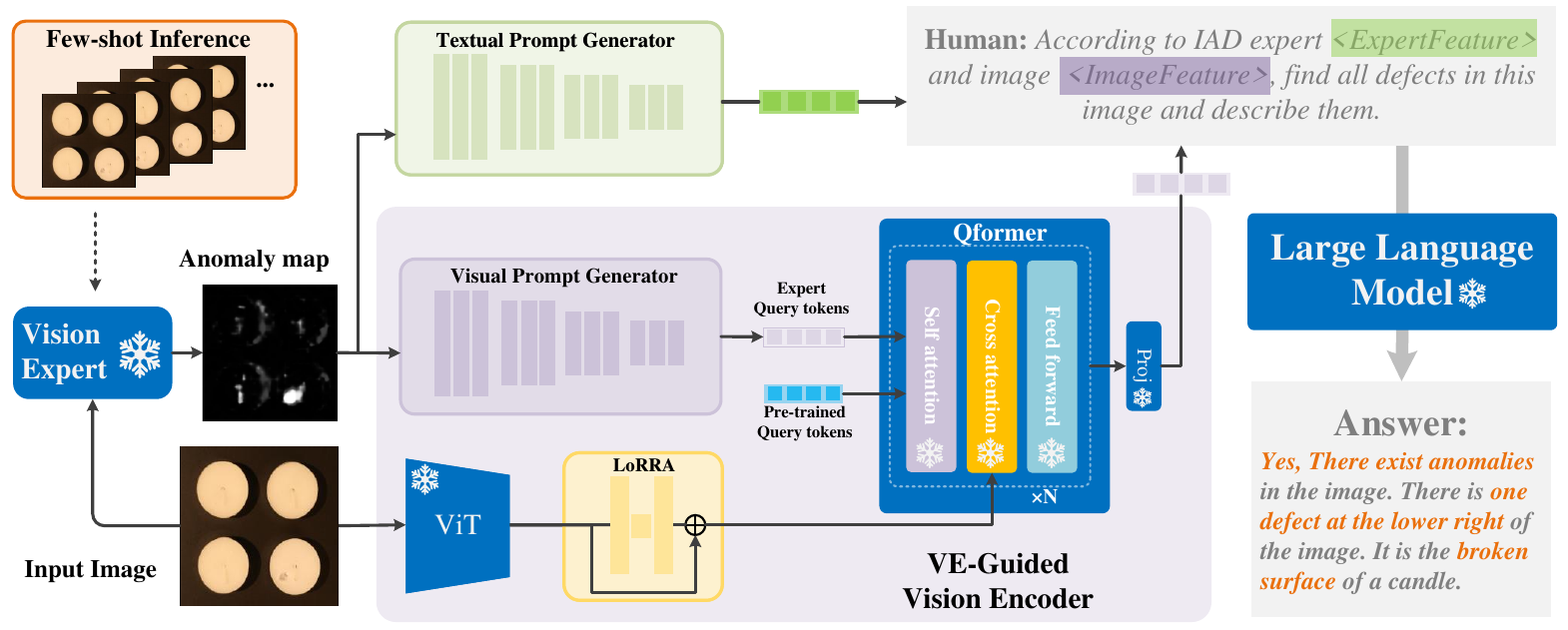}
    \caption{The architecture of proposed Myriad. Given an input industrial image, the vision expert estimates an anomaly map $\mathbf{M}$ containing prior knowledge. To adapt visual features for the IAD task, we propose a VE-guided vision encoder, which enhances vision features for better alignment with industrial images and focuses more on regions via expert prompts generated from the visual prompt generator~(VPG). Furthermore, the textual prompt generator~(TPG) embeds the anomaly map into vision expert tokens, enhancing the LLM's ability to utilize additional information. }
    \label{fig:framework}
    \vspace{-1em}
\end{figure*}

\section{Related Work}

\subsection{Industrial Anomaly Detection}
Given industrial RGB or gray-scale images, IAD methods aim to determine potential anomalies and their location.

Feature embedding-based methods~\cite{SPADE,PaDiM,patchcore,simplenet,DeSTSeg} have made great progress recently. Some methods~\cite{SPADE,PaDiM,patchcore} extract the features of normal samples with pre-trained models~\cite{vggnet,efficientnet,ResNet} and construct a memory bank which is used for comparison to detect anomalies during the inference phase. 
Under the guidance of pre-trained models, some student networks~\cite{DeSTSeg,Deng_2022_CVPR,Distillation} can also be learned to represent the sample space, and anomalies are then predicted by comparing characteristic representations between the student networks and the pre-trained networks.
Though effective, pre-trained models may suffer from domain biases when being applied to industrial images. SimpleNet~\cite{simplenet} addresses this issue by mapping the pre-trained feature space to a domain-specific feature space using a fully connected layer. Additionally, a binary discriminator is trained to detect outliers.

Apart from feature embedding-based methods, there have been other explorations toward more general and effective industrial anomaly detection in recent years.
To eliminate the cost of training individual models for each anomaly type~\cite{patchcore,simplenet,lafite,DiffAD}, some methods~\cite{UniAD,yao2023focus,OmniAL} propose using a unified model to achieve multiple class anomaly detection.
Reconstruction-based methods~\cite{diffusionAD,lafite,DiffAD,removingAD} aim to reconstruct normal images from samples in the training phase. In the inference phase, the anomaly maps can be predicted by comparing original and reconstructed images pixel by pixel.
Language-guided methods~\cite{winclip,aprilgan,AnoVL} leverage the domain alignment abilities of pre-trained multimodal models, such as CLIP~\cite{CLIP}, Region CLIP~\cite{regionclip}, and Imagebind~\cite{imagebind}.
For example, WinCLIP~\cite{winclip} and AprilGAN~\cite{aprilgan} construct two sets of linguistic prompts for abnormal and normal samples, respectively.
Given an industrial image, they extract visual features of the image, which are compared against the textual features of the normal and abnormal prompts at each spatial position.
Since the visual and textual domains are well aligned, such comparison yields the pixel-wise anomaly scores.
AnoVL~\cite{AnoVL} further designs test-time adaptation (TTA) to refine features and improve anomaly location.

However, these IAD methods can only generate image- or pixel-level anomaly scores without determinant detection and detailed descriptions, and they lack the flexibility to generate outputs following the instructions.
To solve these issues, we propose incorporating large vision-language models for industrial anomaly detection.

\subsection{Large Multimodal Models}
Motivated by the impressive cognitive abilities exhibited by LMMs~\cite{blip2,kosmos2,minigpt4,shikra,regionblip,llava,All-Seeing,visionllm}, researchers have embarked on investigating ways for transferring these capacities to the realm of visual perception.
In this context, BLIP-2~\cite{blip2} proposes the utilization of an image encoder to encode visual features, which are fed into LLMs alongside text prompts. Building upon this foundation, LLaVA~\cite{llava} and Mini-GPT4~\cite{minigpt4} initially prioritize establishing alignment between image and text features, followed by in-context instruction tuning. Expanding on these models, Shikra~\cite{shikra} and Kosmos-2~\cite{kosmos2} further enhance the grounding capabilities by referencing objects or regions of interest using definite coordinates or specialized tokens, as opposed to providing detailed textual descriptions.
Nevertheless, the application of these LMMs in industrial anomaly detection encounters challenges stemming from the absence of domain-specific knowledge.
AnomalyGPT~\cite{adgpt} designs a language-guided image decoder to generate an anomaly map, and it employs a prompt learner to learn an expert prompt for LLM. From our perspective, it fails to fully utilize the vision comprehension capacity of existing LMMs for IAD domain images.
By contrast, our work introduces a VE-guided vision encoder, which consists of a visual prompt generator and a low-rank residual adapter to enhance the visual representation by extracting IAD-specific knowledge.
Meanwhile, this work further introduces a textual prompt generator for offering extra information beyond the appearance features into LLM, which includes a binary mechanism to utilize the one-class, zero-shot, and few-shot vision experts during inference.

\begin{table*}[]
    \footnotesize
    \centering
    \tabcolsep 49pt
    \caption{Comparison between LMM-based and existing other methods of industrial anomaly detection across various functionalities.}
    \begin{tabular*}{\textwidth}{@{}ccccc@{}}
        \toprule
        Method & Inference Efficiency & Flexibility & instruction-following & Analysis and Feedback \\
        \midrule
        AOI & Fast & bv Manual Efforts &   \ding{55}   &  Pre-defined items    \\
        Learning-based & Medium  &  by Model Tuning / Memory Update      &    \ding{55} &  \ding{55}    \\
        LMM-based &   Slow      & Ready &   \ding{51}   &   \ding{51}   \\
        \bottomrule
    \end{tabular*}

    \label{tab:comparison}
\end{table*}

\section{Method}

In this section, we first discuss the paradigm of anomaly detection in~\cref{sec:discussion}, then introduce the architecture of Myriad in~\cref{sec:arch}, which consists of the vision expert-guided vision encoder~(\cref{sec:veguided}) and the vision expert guidance adapter~(\cref{sec:veguidance}). Finally, the learning objective and details about IAD instruction are presented in~\cref{sec:learning_objective,sec:data_prepare}.

\subsection{Discussion on Anomaly Detection Paradigm}
\label{sec:discussion}
\noindent\textbf{Automated Optical Inspection (AOI).}
Over the past decades, automated optical inspection (AOI) has been the dominant technique for industrial anomaly detection.
With tens of years of development, the AOI systems are highly optimized and efficient, which are suitable for mass production in modern industry.
However, for a different product, even when only minor changes are applied to the product, the hyper-parameters and configurations should be manually adjusted, making it extremely cumbersome and expensive, and they can only inspect anomaly types with pre-defined rules.
As a result, the AOI systems are typically deployed to high-end manufacturing, and they are gradually struggling to satisfy the requirements of the rapid product iteration and constantly changing demands.

\noindent\textbf{Learning-based Methods.}
Therefore, recent efforts have been directed toward learning-based methods to leverage the power of data-driven optimization schemes, and two main technical routes are applied in the literature.
The first category is to perform IAD directly, where the model predicts the results solely from the test sample~\cite{simplenet,aprilgan} or compares the features of certain normal samples against the test sample~\cite{PaDiM,patchcore,visa,aprilgan}.
The second category reconstructs a ``normal'' counterpart from the test sample via generative models for comparison~\cite{diffusionAD,UniAD,DiffAD,lafite}.
Nonetheless, as shown in \cref{tab:comparison}, the abilities of these methods (\eg, flexibility and instruction-following capability) are still unsatisfactory.

\vspace{0.5em}
\noindent\textbf{Large Multimodal Model-based Method.}
In real-world deployment scenarios, the demands constantly change due to product iteration, design updates, standard changes, and \etc.
To satisfy these requirements, AOI and existing learning-based methods require extensive manual efforts and model tuning, as shown in \cref{tab:comparison}.
Furthermore, the requirements and queries may be open-ended, which may not be achieved by simply adjusting the discrimination principles or fine-tuning the models.
To this end, we propose leveraging the eminent perception and comprehension abilities of LMMs.

\begin{figure*}[t]
    \small
    \centering
    \renewcommand\tabcolsep{2pt}
    \includegraphics[width=0.95\linewidth]{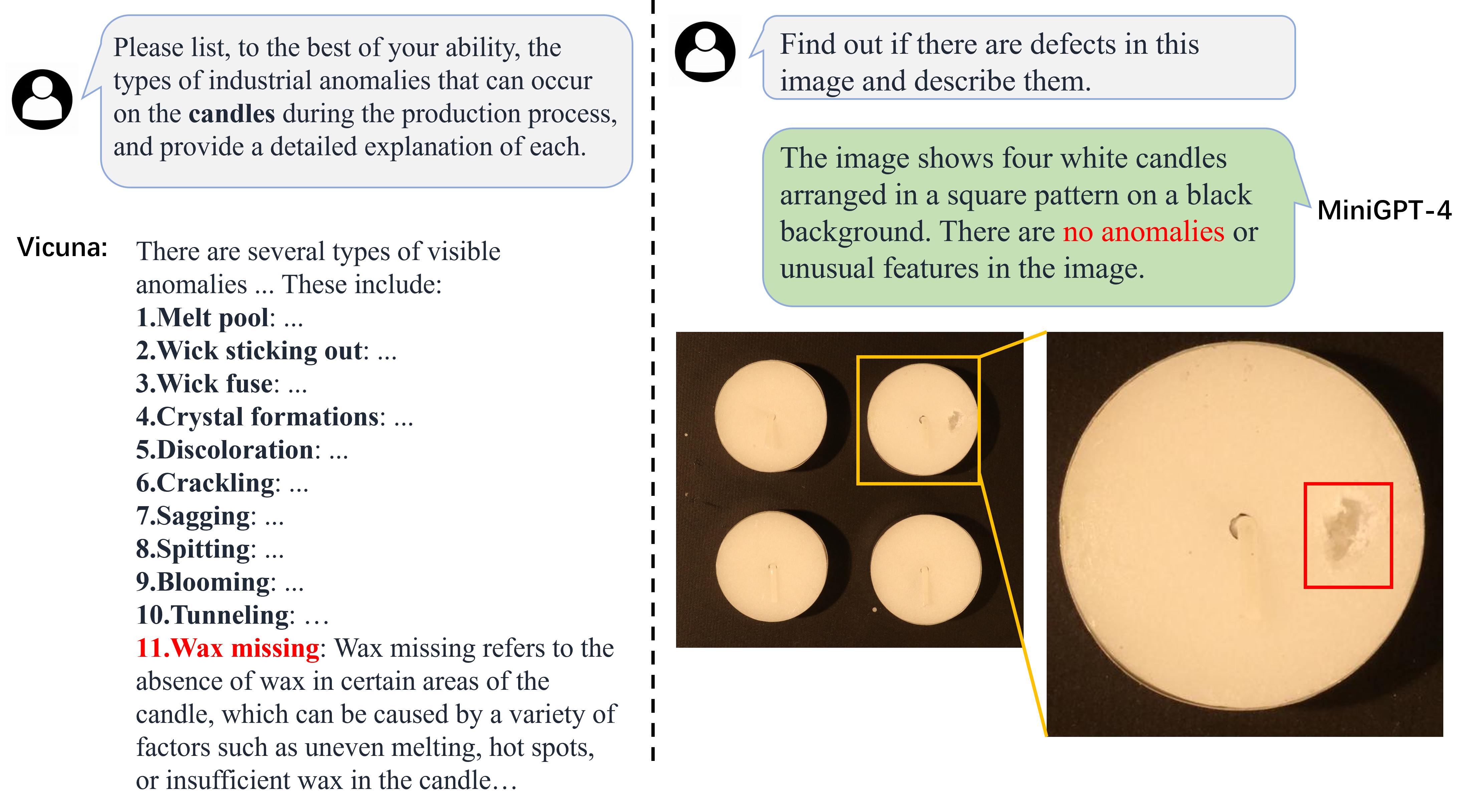}
    \caption{MiniGPT4~\cite{minigpt4} fails to utilize the IAD knowledge in Vicuna~\cite{vicuna} to recognize missing wax on the candles.}
    \label{fig:vicuna-candle}
    \vspace{-1em}
\end{figure*}

As further shown in \cref{fig:vicuna-candle}, we find that even if not specifically designed for anomaly detection, the pre-trained language models utilized in the LMMs can still obtain sufficient professional knowledge about anomaly detection from the large-scale training corpora, including the causes, characteristics, and impacts of the industrial anomalies.
However, such knowledge exists in the language modality only, and the general-purposed LMMs are unable to associate the knowledge with the vision modality due to the limited pairwise vision-language training data for IAD.
In this paper, we identify the lack of specialized IAD-related knowledge as the main factor hindering the application of LMMs for IAD, and propose addressing this issue by introducing existing IAD models as vision experts and fusing the vision expert guidance into LMMs.

\vspace{0.5em}
\subsection{Model Architecture}
\label{sec:arch}
To begin with, we build the LMM framework following MiniGPT-4~\cite{minigpt4}, which comprises two modalities (\ie, vision and language).
For the vision modality, a pre-trained ViT backbone is taken from EVA-CLIP~\cite{evaclip}, followed by the query transformer (Q-Former) from the pre-trained BLIP-2~\cite{blip2}.
While for the language modality, we directly use the pre-trained Vicuna~\cite{vicuna} model.
As analyzed in \cref{sec:discussion}, these models are trained from general-domain training data, and the LMM framework they combine lacks IAD-related knowledge, especially in the vision modality.

Considering that the language model holds a wealth of IAD-related knowledge, to enhance the IAD capability of the LMM framework, we focus on adjusting the vision modality and the vision-to-language knowledge transmission.
For the former, we leverage existing IAD methods as vision experts, whose outputs (\ie, the anomaly map) are utilized to guide the feature extraction from the industrial image $\mathbf{I}$ in the vision modality.
The proposed VE-guided vision encoder not only extracts IAD-suitable visual features (denoted by $\mathbf{t}_\mathit{v}$) under the guidance of vision experts, but can also learn from anomaly maps that contain the underlying cognition and comprehension of IAD models.
For the latter, the anomaly map is fed into the language model as another input (denoted $\mathbf{t}_\mathit{e}$), hoping that it can deliver additional information to the language model (\eg, positional information), as the vision encoder usually extracts more visual appearance features.

In this way, given a text instruction $\mathbf{t}_\mathit{i}$, Myriad is capable of generating text response $\mathbf{R}$ corresponding to the instruction such as the anomaly detection results and detailed descriptions about the anomalies, \ie,
\begin{equation}
    \mathbf{R}=\mathit{LLM}(\mathbf{t}_i,\mathbf{t}_\mathit{v},\mathbf{t}_\mathit{e}),
\end{equation}
where both $\mathbf{t}_\mathit{e}$ and $\mathbf{t}_\mathit{v}$ are obtained from $\mathbf{I}$, and $\mathit{LLM}$ denotes an large language model.

\subsection{Vision Expert-Guided Vision Encoder}
\label{sec:veguided}

To obtain an IAD-suitable $\mathbf{t}_\mathit{v}$ from $\mathbf{I}$, we design the vision expert-guided vision encoder as follows.
First, we let the vision encoder take anomaly maps as input as well, by introducing a visual prompt generator that produces expert query tokens $\mathbf{q}_\mathit{e}$ tailored to the original Q-former query tokens, \ie,
\begin{equation}
    \mathbf{q}_\mathit{e}=G_{Q} ( VE(\mathbf{I});\theta_\mathit{G_{Q}}),
\end{equation}
where $G_{Q}$ and $\theta_{\mathit{G_Q}}$ denote the visual prompt generator for Q-former and its parameters, respectively.
Second, we introduce a trainable low-rank residual adapter $\mathit{A}$~ (LoRRA) that is cascaded to the pre-trained ViT to generate visual representations (denoted by $\mathbf{v}_\mathit{e}$) that are modulated for the IAD task. To avoid over-fitting to a certain dataset, the architecture takes inspiration from LoRA~\cite{lora} which optimizes rank decomposition weight matrices instead of dense layers. The proposed LoRRA consists of two convolutional layers. The first layer compresses the visual features to four dimensions, and the second layer remaps them back to their original dimensionality as part of a residual connection.
\begin{equation}
 \mathbf{v}_\mathit{e}=\mathbf{v}_{\mathbf{I}}+\mathit{A}(\mathbf{v}_{\mathbf{I}}),
\end{equation}
Note that $\mathbf{v}_{\mathbf{I}}$ is the visual feature extracted from the image $\mathbf{I}$ through pre-trained ViT. 
Together with the pre-trained query tokens $\mathbf{q}_{0}$ in Q-former, the expert query tokens $\mathbf{q}_\mathit{e}$ then jointly interact with the adapted visual representations of industrial images (obtained from the ViT and the visual adapter) through cross-attention, so as to generate visual representations $\mathbf{f}_\mathit{p}$ following prior knowledge from the experts, \ie,
\begin{equation}
    \mathbf{f}_\mathit{p}=\mathit{Q}(\mathbf{q}_{0},\mathbf{q}_\mathit{e};\mathbf{v}_\mathit{e}),
\end{equation}
With $\mathbf{f}_\mathit{p}$, finally, we use the pre-trained projection layer $\mathit{MLP}$ in MiniGPT-4 to obtain the visual representations that are 1) suitable to the IAD task and 2) understandable by the LLM:
\begin{equation}
    \mathbf{t}_\mathit{v}=\mathit{MLP}(\mathbf{f}_\mathit{p}),
    \vspace{-2mm}
\end{equation}

With these designs, Myriad not only performs accurate IAD with high-quality anomaly maps from the vision experts but also achieves favorable performance when the vision experts fail. See evidence in \cref{sec:qualitative}.

\subsection{Prompt Generator}
\label{sec:veguidance}
Traditional IAD models provide anomaly maps, which contain sufficient prior knowledge of the IAD task. To enable LLMs to receive extra auxiliary information such as positional prior, we design the prompt generator, which aims to transform the anomaly maps into prompts $\mathbf{t}_e$ that compress priors from IAD vision experts.
\begin{equation}
    \mathbf{p}_\mathit{*}=G_{*}(\mathbf{M};\theta_\mathit{G_{*}}),
\end{equation}
where $G_{*}$ and $\theta_{\mathit{G_{*}}}$ are denoted as the prompt generator and its parameters, respectively, and $\mathbf{M} \in \mathbb{R}^{H \times W} $ represents the anomaly map of $\mathbf{I}$ produced by certain vision expert. $\mathbf{p}_\mathit{*}$ is denoted as $\mathbf{p}_\mathit{LLM}$ or $\mathbf{p}_\mathit{Q}$.
Note that in Myriad, we use two prompt generators~($G_{T}$ and $G_{V}$) to prompt LLM and Q-former respectively.
As shown in \cref{fig:framework}, the prompt generator contains several blocks, each consisting of a convolution with 3$\times$3 kernel, a ReLU as the activation function, and a max pooling, to map its input anomaly map $\mathbf{M}$ into expert prompts $\mathbf{p}_\mathit{*} \in \mathbb{R}^{D_{VE} \times D_{in,*}} $, where $ D_{in,*} $ is the input dimension of the target module~(LLM or Q-former) and $ D_{VE} $ is the number of the expert prompt. In our experiments, $ D_{VE}$ is set to $9$ for LLM and $49$ for Q-former.

Since the prompt generator only takes anomaly maps as inputs, it makes the vision expert tokens decoupled from the appearance and domain of colorful images to some extent, which can be more robust to domain shift in practice.

\subsection{Learning Objective}
\label{sec:learning_objective}
The training of our Myriad contains two stages. 
At the first stage, we pre-train the weights of VE-guided vision encoder and textual prompt generator for LLM separately, such that $\mathbf{t}_\mathit{v}$ and $\mathbf{p}_\mathit{LLM}$~(also denoted as $\mathbf{t}_\mathit{e}$) contain reliable representations and also can be well received by LLM, \ie,
\begin{equation}\label{eq:7}
    \mathcal{L}_{ce}=-\sum^{n}_{k=1} y_k\log(\mathit{LLM}(\mathbf{t}_{i},\mathbf{t}_{*})),
\end{equation}
where $y_k$ is given to represent the target text sequence. $\mathbf{t}_{*}$ is denoted as $\mathbf{t}_\mathit{e}$ or $\mathbf{t}_\mathit{v}$. Cross-entropy loss $\mathcal{L}_{CE}$ is commonly employed for training language models.
At the end of the stage, we obtain pre-trained weights of both textual prompt generator and VE-guided vision encoder. 

At the second stage, the pre-trained weights of textual prompt generator and VE-guided vision encoder are loaded and fine-tuned as shown in~\cref{fig:framework}. Myriad are activated to make use of both visual information guided by experts and direct auxiliary hints from the vision expert outputs.
Formally, we jointly train all parameters with combined losses:
\begin{equation}
    \mathcal{L}_{ce}=-\sum^{n}_{k=1} y_k\log(\mathit{LLM}(\mathbf{t}_i,\mathbf{t}_{v},\mathbf{t}_e)),
    \hspace{-2mm}
\end{equation}

\subsection{IAD Instruction Data Preparation}
\label{sec:data_prepare}

Due to the lack of detailed annotations and the small amount of IAD datasets, we train Myriad with only binary anomaly detection tasks. To construct the task, we design several instruction templates. The templates are given by paraphrasing, for example, ``\textit{According to IAD expert $<$Img$>$ $<$ExpertFeature$>$ $<$/Img$>$ and image $<$Img$>$ $<$ImageFeature$>$ $<$/Img$>$, find all defects in this image.}'', where \textit{$<$ImageFeature$>$} is a placeholder of the visual tokens produced from the VE-guided vision encoder and \textit{$<$ExpertFeature$>$} is another placeholder for prompts extracted from the textual prompt generator.
While pre-training textual prompt generator and VE-guided vision encoder separately, the instruction templates are ``\textit{According to IAD expert $<$Img$>$ $<$ExpertFeature$>$ $<$/Img$>$, find all defects in this image.}'' and ``\textit{According to the image $<$Img$>$ $<$ImageFeature$>$ $<$/Img$>$, find all defects in this image}'', respectively. 
For the ground truth response, if anomalies are detected, the response is ``Yes, anomalies exist in this image.'' Otherwise, it is ``No, there are no anomalies in this image.''

To produce vision-language pairs in one-class setting where only normal images are available, we follow the similar anomaly simulation methods in the previous work~\cite{NSA, adgpt} to generate synthetic anomalies from normal images.
More specifically, the synthetic anomalies are generated using the cut-paste method, where several random regions in the source images are cut and pasted to a random position in each target image.

\section{Experiments}

\noindent\textbf{Datasets.}
Our experiments are performed on the MVTec-AD~\cite{mvtec} and VisA dataset~\cite{visa}.
The MVTec-AD dataset~\cite{mvtec} consists of $3,629$ samples in the train set and $1,725$ samples in the test set, respectively.
MVTec-AD contains $15$ sub-datasets across different types of industrial products, including 5 textual sub-datasets and 10 object sub-datasets, making it the classical dataset in industrial anomaly detection.
The VisA dataset~\cite{visa} contains $9,621$ normal and $1,200$ anomalous samples, and covers $12$ objects. VisA has multiple objects, a complex background, and fewer abnormal samples, making it a dataset closer to actual industrial applications than MVTec-AD. Therefore, VisA has become a popular and more challenging benchmark.
For both MVTec-AD and VisA, we train Myriad following standard one-class data split setting~\cite{visa}, where only normal samples are visible during training, and all abnormal samples are used for testing.

\vspace{0.5em}
\noindent\textbf{Evaluation Settings.}
This work performs experiments on three settings: one-class setting, zero-shot setting, and few-shot setting. 
For one-class setting, models are allowed to train with normal images of products. 
For zero-shot setting, products are unseen for the IAD models. 
For few-shot setting, ``$N$-shot'' means that $N$ normal images of a product are visible during inference.

\vspace{0.5em}
\noindent\textbf{Evaluation Metrics.}
By applying the vision experts, Myriad can also provide an anomaly map, thus we can first follow previous works to report its I-AUROC and P-AUROC performance with the experts, although (considering that our Myriad utilizes previous vision experts to obtain anomaly maps) it is easy to match its I-AUROC and P-AUROC performance to previous state-of-the-arts.
In addition, we further report the mean accuracy across products based on the text response of LMMs as a main metric for LMM-based methods. 
The threshold used to compute accuracy for conventional IAD methods is determined with the max scores of normal samples over $k$-fold evaluations. Note that the process is only performed on the train set.
In practice, we set $k=3$. 

\begin{table}[]
    \centering
    \small
    \renewcommand\arraystretch{1.2}
    \caption{One-class anomaly detection results on MVTec-AD dataset. The best-performing method is in \textbf{bold}.}
    \begin{tabular}{@{}cccc@{}}
        \toprule
        Method                    & Image-AUC                  & Pixel-AUC                  & Accuracy                \\ \midrule
        PaDiM                     & 95.3            & 97.4             & 76.5          \\
        PatchCore                 & 99.0            & \textbf{98.1}    & 89.2          \\
        SimpleNet                 & \textbf{99.6}    & 98.1            & 93.0          \\
        UniAD                     & 97.6                       & 97.0                       & 89.3                    \\
        AnomalyGPT                & 97.4                       & 93.1                       & 93.3                    \\ \midrule
        \textbf{Myriad~(ours)}  & 97.4                       & 93.1                       & \textbf{94.2}           \\ \bottomrule
    \end{tabular}\vskip -0.1in

    \label{tab:unsupervised}
\end{table}

\vspace{0.5em}
\noindent\textbf{Implementation Details.}
We utilize MiniGPT-4 as the base LMM and test in three different settings: the zero-shot setting, the few-shot setting, and the typical one-class setting.
In each setting, we use different vision experts.
In the zero-shot setting, AprilGAN~\cite{aprilgan} is used as a baseline. We introduce MuSc~\cite{Li2024MuSc} to evaluate the plug-and-play capability of the prompt generators and to serve as a more effective expert for improving Myriad’s anomaly detection performance during inference.
In the classical one-class settings, we use the image decoder and ImageBind backbone~\cite{imagebind} in AnomalyGPT~\cite{adgpt} as the vision expert.
In the few-shot setting, the anomaly maps are obtained in the same way as Patchcore~\cite{patchcore}, by adapting ImageBind~\cite{imagebind} as the backbone and estimating cosine similarity between the features of the test images and those of the few-shot normal samples.

Our training strategy consists of a pre-training and a fine-tuning stage. In the pre-training stage, we train the vision-expert guidance adapter and the vision-expert guided visual feature extraction module independently. Both two modules would be combined and fine-tuned in the fine-tuning stage with a lower learning rate and real abnormal data.
The pre-training process lasts for 32,000 steps with a batch size of 16. In each training batch, half of the samples are real normal samples, and the other half contains simulated abnormal samples~\cite{NSA}. AdamW, with a weight decay of 0.05, is employed as the optimizer. No warmup step is used. The cosine learning rate decay strategy is applied, and the minimum learning rate is $0$.
In the fine-tuning phase, the training process lasts for 16,000 steps with a batch size of 16. AdamW, with a learning rate of 3e-5, is performed for fine-tuning. The cosine learning rate decay strategy is also used, and the final learning rate is set to 1e-5.
All experiments are conducted on 2 NVIDIA A100 GPUs.

\subsection{Quantitative Results}
\label{sec:main_results}

In this section, we report quantitative comparison results between Myriad and previous IAD works, including feature-embedding-based methods, reconstruction-based methods, and language-guided methods.
Given different vision experts, Myriad can perform as zero-shot, few-shot, and one-class anomaly detectors.

\vspace{0.5em}
\noindent\textbf{One-class IAD.}
We report one-class IAD results on MVTec-AD in~\cref{tab:unsupervised}.
Myriad achieves superior accuracy than the state-of-the-art methods on MVTec-AD. Our solution outperforms PatchCore by about $5.0$\% ($94.2$\% vs $89.2$\%) and outperforms SimpleNet by about $1.2$\% ($94.2$\% vs $93.0$\%).
Moreover, with the same anomaly maps predicted by AnomalyGPT, Myriad outperforms AnomalyGPT by about $0.9$\% ($94.2$\% vs $93.3$\%). 
To ensure a fair comparison, we use the same training data and identical prompt-generator architecture for the LLM as in the prompt learner of AnomalyGPT.
These results indicate that the VE-guided vision encoder extracts superior IAD visual features using the 0.4-billion-parameter EVA-CLIP model compared to the single-class image token produced by the 1.2-billion-parameter backbone employed in AnomalyGPT.

\begin{table*}[]
    \centering
    \caption{Zero/Few-shot IAD results on MVTec-AD and VisA datasets. Results are listed as the average of 5 runs and the best-performing method is in \textbf{bold}. The results for SPADE and WinCLIP are reported from~\cite{winclip}. The models are cross-validated across two datasets: MVTec-AD~\cite{mvtec} and VisA~\cite{visa}.}
    \resizebox{\linewidth}{!}{
    \begin{tabular}{@{}cccccccc@{}}
        \toprule
        \multirow{2}{*}{Setup} &
        \multirow{2}{*}{Method} &
        \multicolumn{3}{c}{MVTec-AD} &
        \multicolumn{3}{c}{VisA} \\ \cmidrule(lr){3-5} \cmidrule(lr){6-8}
        &&Image-AUC &Pixel-AUC &Accuracy &Image-AUC &Pixel-AUC &Accuracy \\ \midrule
        \multirow{4}{*}{0-shot}
        &MuSc & 97.8~\stdvu{$\pm$ 0.1}  & 97.1~\stdvu{$\pm$ 0.1}  & -  & 92.6~\stdvu{$\pm$ 0.2}  & 98.7~\stdvu{$\pm$ 0.2}  & - \\
        &{Myriad~(MuSc)} & 97.8~\stdvu{$\pm$ 0.1}  & 97.1~\stdvu{$\pm$ 0.1}  & \textbf{86.8~\stdvu{$\pm$ 0.1}}& 92.6~\stdvu{$\pm$ 0.2} & 97.3~\stdvu{$\pm$ 0.2} & 78.7~\stdvu{$\pm$ 0.2} \\ \cmidrule(l){2-8}
        &AprilGAN & 86.2~\stdvu{$\pm$ 0.5} & 87.6~\stdvu{$\pm$ 0.1} & 69.3~\stdvu{$\pm$ 0.4} & 78.0~\stdvu{$\pm$ 0.4} & 94.2~\stdvu{$\pm$ 0.3} & 61.8~\stdvu{$\pm$ 0.2} \\
        &\textbf{Myriad~(AprilGAN)} &86.2~\stdvu{$\pm$ 0.5} &87.6~\stdvu{$\pm$ 0.1} &\textbf{72.2~\stdvu{$\pm$ 0.2} } & 78.0~\stdvu{$\pm$ 0.4} & 94.2~\stdvu{$\pm$ 0.3} & \textbf{63.2~\stdvu{$\pm$ 0.2}} \\ \midrule
        \multirow{8}{*}{1-shot}
        &SPADE &81.0~\stdvu{$\pm$ 2.0} &91.2~\stdvu{$\pm$ 0.4} &- &79.5~\stdvu{$\pm$ 4.0} &95.6~\stdvu{$\pm$ 0.4} &- \\
        &PaDiM &75.5~\stdvu{$\pm$ 1.0} &90.0~\stdvu{$\pm$ 0.4} &57.4~\stdvu{$\pm$ 1.6} &59.6~\stdvu{$\pm$ 1.8} &91.3~\stdvu{$\pm$ 0.3} &45.0~\stdvu{$\pm$ 0.6} \\
        &PatchCore &84.2~\stdvu{$\pm$ 1.2} &92.4~\stdvu{$\pm$ 0.5} &65.0~\stdvu{$\pm$ 1.8} &76.8~\stdvu{$\pm$ 1.6} &93.5~\stdvu{$\pm$ 0.6} &60.0~\stdvu{$\pm$ 1.0} \\
        &WinCLIP &93.1~\stdvu{$\pm$ 2.0} &95.2~\stdvu{$\pm$ 0.5} &- &83.8~\stdvu{$\pm$ 4.0} &\textbf{96.4~\stdvu{$\pm$ 0.4}} &- \\
        &AprilGAN &{92.0~\stdvu{$\pm$ 0.3}} &{95.1~\stdvu{$\pm$ 0.1}} &{75.6~\stdvu{$\pm$ 0.1}} & \textbf{91.2~\stdvu{$\pm$ 0.8}} &96.0~\stdvu{$\pm$ 0.0} &{70.0~\stdvu{$\pm$ 0.2}} \\
        &AnomalyGPT &\textbf{94.1~\stdvu{$\pm$ 1.1}} &\textbf{95.3~\stdvu{$\pm$ 0.1}} &{86.1~\stdvu{$\pm$ 1.1}} &\textbf{87.4~\stdvu{$\pm$ 0.8}} &96.2~\stdvu{$\pm$ 0.1} &{77.4~\stdvu{$\pm$ 1.0}} \\ \cmidrule(l){2-8}
        &{Myriad~(AprilGAN)} &92.0~\stdvu{$\pm$ 0.3} &{95.1~\stdvu{$\pm$ 0.1}} & 76.5~\stdvu{$\pm$ 0.3} &{91.2~\stdvu{$\pm$ 0.8}} &96.0~\stdvu{$\pm$ 0.0} &{72.7~\stdvu{$\pm$ 0.1}} \\
        &\textbf{Myriad~(AnomalyGPT)} &\textbf{94.1~\stdvu{$\pm$ 1.1}} &\textbf{95.3~\stdvu{$\pm$ 0.1}} &\textbf{87.4~\stdvu{$\pm$ 0.9 }} &\textbf{87.4~\stdvu{$\pm$ 0.8}} &96.2~\stdvu{$\pm$ 0.1} &\textbf{80.0~\stdvu{$\pm$ 0.4}} \\ \midrule
        \multirow{8}{*}{2-shot}
        &SPADE &82.9~\stdvu{$\pm$ 2.6} &92.0~\stdvu{$\pm$ 0.3} &- &80.7~\stdvu{$\pm$ 5.0} &96.2~\stdvu{$\pm$ 0.4} &- \\
        &PaDiM &78.2~\stdvu{$\pm$ 0.6} &92.1~\stdvu{$\pm$ 0.4} &56.9~\stdvu{$\pm$ 0.8} &65.5~\stdvu{$\pm$ 1.5} &93.2~\stdvu{$\pm$ 0.1} &46.4~\stdvu{$\pm$ 0.7} \\
        &PatchCore &87.1~\stdvu{$\pm$ 0.8} &94.1~\stdvu{$\pm$ 0.2} &68.4~\stdvu{$\pm$ 2.3} &80.4~\stdvu{$\pm$ 0.7} &95.0~\stdvu{$\pm$ 0.2} &61.8~\stdvu{$\pm$ 1.2} \\
        &WinCLIP &94.4~\stdvu{$\pm$ 1.3} &\textbf{96.0~\stdvu{$\pm$ 0.3}} &- &84.6~\stdvu{$\pm$ 2.4} &\textbf{96.8~\stdvu{$\pm$ 0.3}} &- \\
        &AprilGAN &{92.4~\stdvu{$\pm$ 0.3}} &{95.5~\stdvu{$\pm$ 0.0}} &{76.0~\stdvu{$\pm$ 0.2}} &\textbf{92.2~\stdvu{$\pm$ 0.3}} &96.2~\stdvu{$\pm$ 0.0} &{71.5~\stdvu{$\pm$ 0.1}} \\
        &{AnomalyGPT} &\textbf{95.5~\stdvu{$\pm$ 0.8}} &95.6~\stdvu{$\pm$ 0.2}&{84.8~\stdvu{$\pm$ 0.8}} &\textbf{88.6~\stdvu{$\pm$ 0.7}} &96.4~\stdvu{$\pm$ 0.1} &{77.5~\stdvu{$\pm$ 0.3}} \\ \cmidrule(l){2-8}
        &{Myriad~(AprilGAN)} &{92.4~\stdvu{$\pm$ 0.3}} &{95.5~\stdvu{$\pm$ 0.}} & 77.1~\stdvu{$\pm$ 0.2} &{92.2~\stdvu{$\pm$ 0.3}} &96.2~\stdvu{$\pm$ 0.0} & 75.3~\stdvu{$\pm$ 0.4} \\
        &\textbf{Myriad~(AnomalyGPT)} &\textbf{95.5~\stdvu{$\pm$ 0.8}} &95.6~\stdvu{$\pm$ 0.2}&\textbf{86.2~\stdvu{$\pm$ 0.7} }&\textbf{88.6~\stdvu{$\pm$ 0.7}} &96.4~\stdvu{$\pm$ 0.1}&\textbf{82.3~\stdvu{$\pm$ 0.5}} \\ \midrule
        \multirow{8}{*}{4-shot}
        &SPADE &84.8~\stdvu{$\pm$ 2.5} &92.7~\stdvu{$\pm$ 0.3} &- &81.7~\stdvu{$\pm$ 3.4} &96.6~\stdvu{$\pm$ 0.3} &- \\
        &PaDiM &80.9~\stdvu{$\pm$ 0.9} &94.0~\stdvu{$\pm$ 0.2} &57.9~\stdvu{$\pm$ 1.2} &69.6~\stdvu{$\pm$ 1.5} &94.4~\stdvu{$\pm$ 0.1} &48.0~\stdvu{$\pm$ 1.4} \\
        &PatchCore &89.5~\stdvu{$\pm$ 1.3} &94.9~\stdvu{$\pm$ 0.2} &72.5~\stdvu{$\pm$ 1.8} &82.2~\stdvu{$\pm$ 0.8} &96.0~\stdvu{$\pm$ 0.1} &63.1~\stdvu{$\pm$ 0.4} \\
        &WinCLIP &95.2~\stdvu{$\pm$ 1.3} &96.2~\stdvu{$\pm$ 0.3} &- &87.3~\stdvu{$\pm$ 1.8} &\textbf{97.2~\stdvu{$\pm$ 0.2}} &- \\
        &AprilGAN &{92.8~\stdvu{$\pm$ 0.2}} &{95.9~\stdvu{$\pm$ 0.0}} &{77.2~\stdvu{$\pm$ 0.1}} & \textbf{92.6~\stdvu{$\pm$ 0.4}} &96.2~\stdvu{$\pm$ 0.0} &{71.6~\stdvu{$\pm$ 0.1}} \\
        &{AnomalyGPT} &\textbf{96.3~\stdvu{$\pm$ 0.3}} &\textbf{96.2~\stdvu{$\pm$ 0.1}} &{85.0~\stdvu{$\pm$ 0.3} }&\textbf{90.6~\stdvu{$\pm$ 0.7}} &96.7~\stdvu{$\pm$ 0.1} &{77.7~\stdvu{$\pm$ 0.4}} \\  \cmidrule(l){2-8}
        &{Myriad~(AprilGAN)} &{92.8~\stdvu{$\pm$ 0.2}} &{95.9~\stdvu{$\pm$ 0.0}} & 77.8~\stdvu{$\pm$ 0.2}&{92.6~\stdvu{$\pm$ 0.4}} &96.2~\stdvu{$\pm$ 0.0} &76.5~\stdvu{$\pm$ 0.2} \\
        &\textbf{Myriad~(AnomalyGPT)} &\textbf{96.3~\stdvu{$\pm$ 0.3}} &\textbf{96.2~\stdvu{$\pm$ 0.1}} &\textbf{86.0~\stdvu{$\pm$ 0.3}} &\textbf{90.6~\stdvu{$\pm$ 0.7}} &96.7~\stdvu{$\pm$ 0.1} & \textbf{83.5~\stdvu{$\pm$ 0.3}} \\ \bottomrule
    \end{tabular}}\vskip -0.1in

    \label{tab:few-shot}\vskip -0.1in
\end{table*}

\vspace{0.5em}
\noindent\textbf{Zero-shot IAD.}
With zero-shot vision experts, such as AprilGAN~\cite{aprilgan} and MuSc~\cite{Li2024MuSc}, Myriad can also perform as a zero-shot anomaly detector.
As demonstrated in~\cref{tab:few-shot}, Myriad with vision expert AprilGAN achieves 2.9\% performance~(72.2\% \vs 69.3\%) on MVTec-AD and 1.4\%~(63.2\% \vs 63.2\%) on VisA.
Without finetuning, MuSc can be used as a plug-and-play vision expert. Myriad with MuSc achieves 86.8\% on MVTec-AD and 78.7\% on VisA. Compared with AprilGAN, Myriad gains 14.6\% and 15.5\% improvement on MVTec-AD and VisA, respectively. The result indicates that better vision experts are able to boost Myriad without any additional cost.

PCB Bank~\cite{pcb_bank} is a novel benchmark for printed circuit board (PCB) anomaly detection. As in~\cref{tab:pcbbank_fewshot}, we apply zero-shot evaluation to its test set and report results across several state-of-the-art general-domain large multimodal models (LMMs) and LMM-based methods. Myriad, based on a 7-billion-parameter LLM, achieves performance comparable to the 72-billion-parameter LLaVA-OneVision model, which is trained on 1.3 million multimodal data samples.

Compared to AnomalyGPT, Myriad improves accuracy by 11.7\% (61.8\% vs. 50.1\%). For a fair comparison, we replace the input anomaly maps for the prompt learner in AnomalyGPT with binarized anomaly maps predicted by MuSc. This modification effectively boosts AnomalyGPT’s accuracy from 50.1\% to 52.3\%. However, Myriad still outperforms AnomalyGPT by 9.5\%.
The performance gap between boosted AnomalyGPT and Myriad is attributed to the vision modality. We notice the detailed nature of PCB Bank, which includes numerous components and multiple labels. 
AnomalyGPT only provides a single class token with pre-trained ImageBind to the LLM, representing a global visual feature that loses many detailed aspects of the PCB. In contrast, Myriad supplies patch tokens and resamples them using the Q-former and expert prompts, resulting in more comprehensive anomaly detection compared to AnomalyGPT.

\vspace{0.5em}
\noindent\textbf{Few-shot IAD.}
In the few-shot setting, we mainly compare with SPADE~\cite{SPADE}, PaDiM~\cite{PaDiM}, PatchCore~\cite{patchcore}, WinCLIP~\cite{winclip}, and AnomalyGPT~\cite{adgpt}.
These methods all have a memory bank but have different matching strategies or augmentations with few references.
We follow~\cite{adgpt} to adopt ImageBind~\cite{imagebind} to generate few-shot anomaly maps by computing cosine similarity features of test images and that of few-shot normal samples. This expert is denoted as AnomalyGPT because it keeps the same I-AUROC and P-AUROC as AnomalyGPT. 

Results are reported in~\cref{tab:few-shot}. 
Myriad combined with AnomalyGPT achieves the highest one-shot accuracy among both conventional IAD methods and LMM-based approaches, attaining 80.0\% on VisA and 87.4\% on MVTec-AD. 
In the two-shot and four-shot settings, Myriad’s performance decreases compared to the one-shot scenario due to internal variations among normal samples. We adopt AnomalyGPT’s selection of few-shot examples, and a similar performance decline is observed in AnomalyGPT. Nevertheless, Myriad still maintains the best accuracy in both two-shot and four-shot settings across MVTec-AD and VisA.

\begin{figure*}
\centering
\begin{subfigure}{.32\linewidth}
    \centering
    \includegraphics[width=0.95\linewidth]{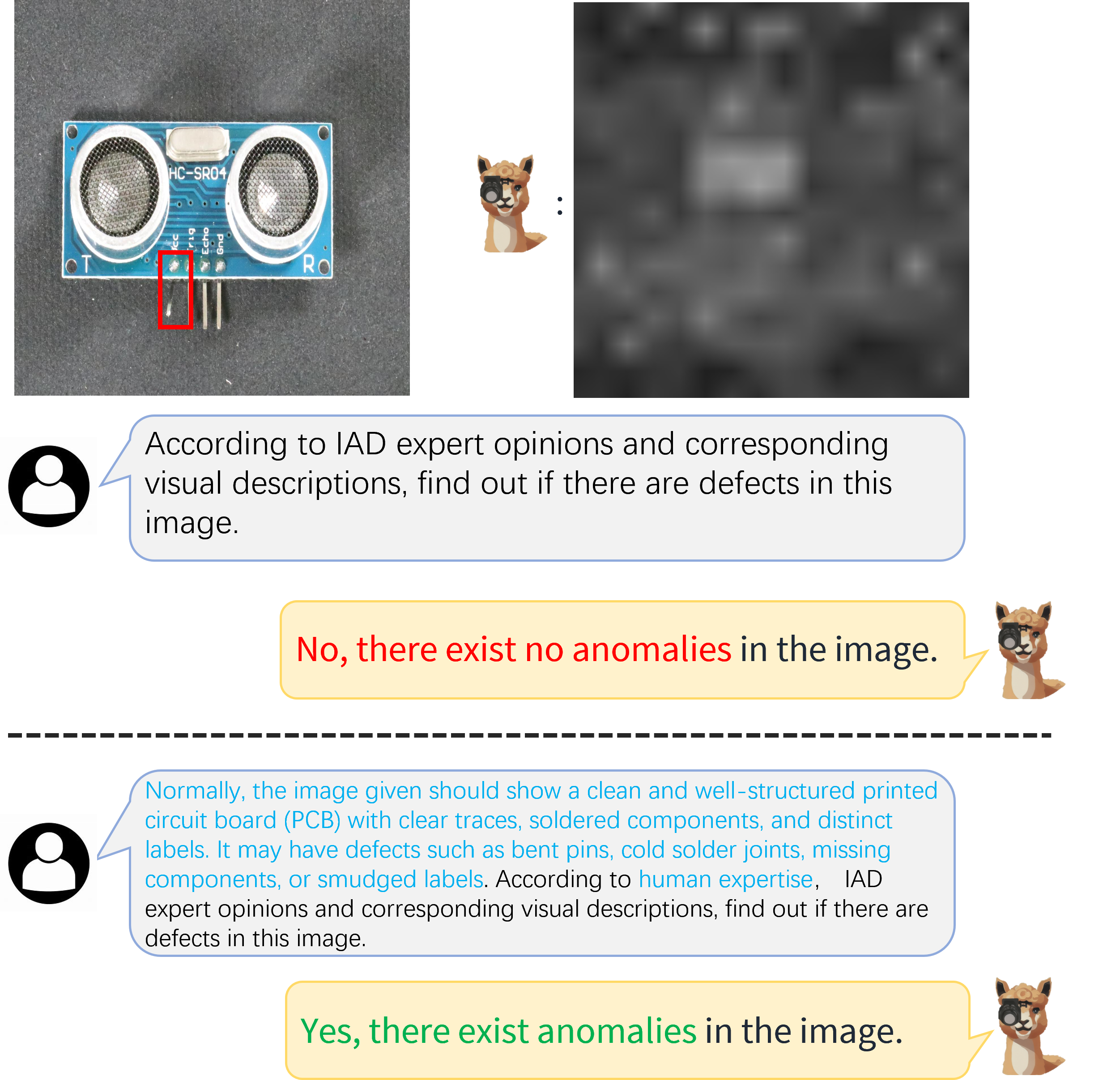}
    \caption{}
    \label{fig:human1}
\end{subfigure}
\begin{subfigure}{.32\linewidth}
    \centering
    \includegraphics[width=0.95\linewidth]{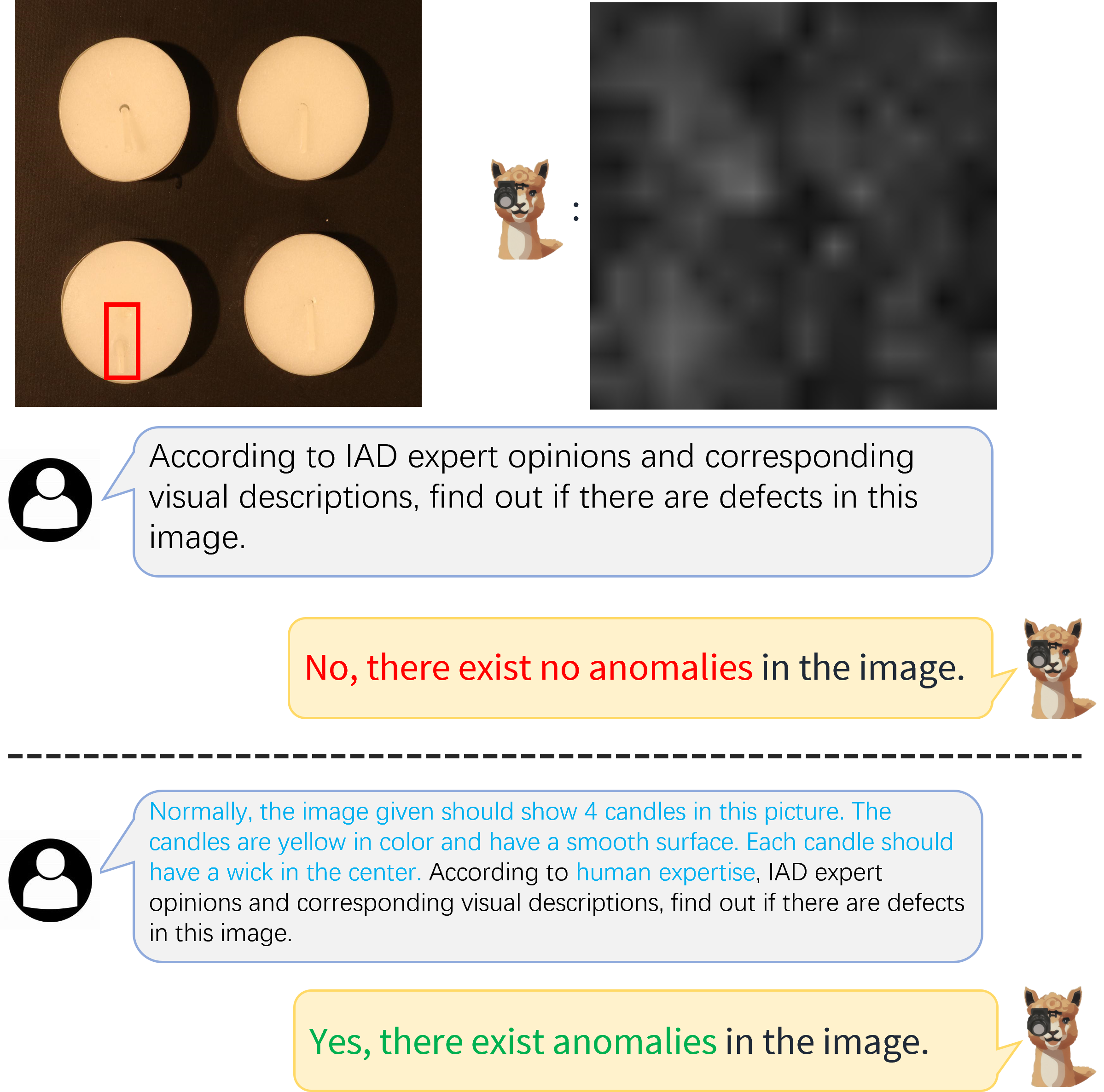}
    \caption{}
    \label{fig:human2}
\end{subfigure}
\begin{subfigure}{.32\linewidth}
    \centering
    \includegraphics[width=0.95\linewidth]{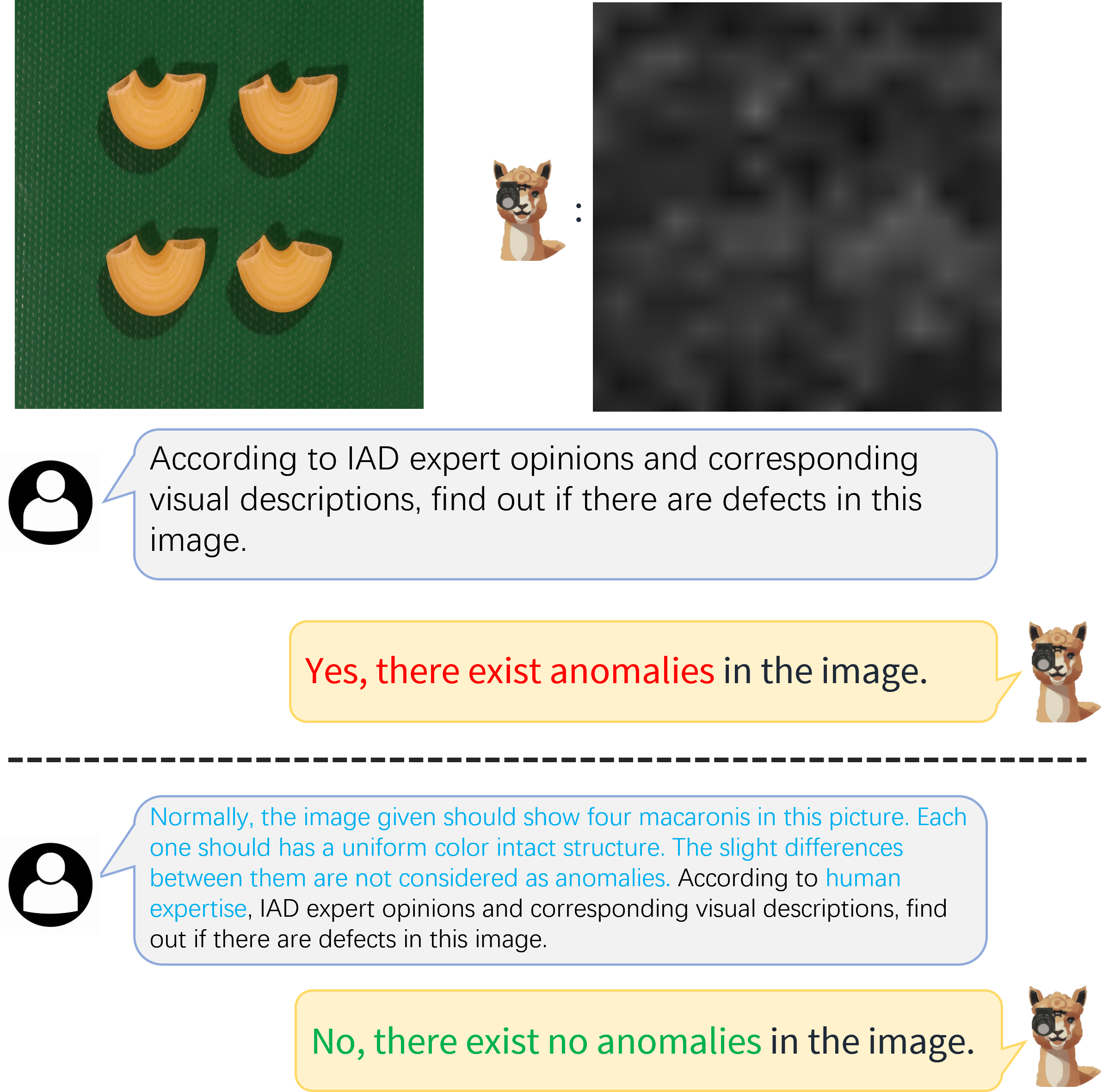}
    \caption{}
    \label{fig:human3}
\end{subfigure}
\caption{\textbf{The illustration of flexibility.} Myriad inherits the flexibility from LMMs. Myriad receives human expertise and
further generates correct text sequence. The ground truth defects are highlighted with \textbf{\color{red} red} bounding boxes in the image.}
\label{fig:normal_hint}
\end{figure*}

\begin{table}[]
    \centering
    \caption{Zero-shot IAD results on PCB-Bank datasets. AnomalyGPT and Myriad are trained on MVTec-AD, because PCB-Bank contains examples of VisA. The best-performing method is in \textbf{bold}}
    \begin{tabular}{@{}cccccc@{}}
        \toprule
        Setup & Method & params & Image-AUC & Pixel-AUC &Accuracy \\ \midrule
        \multirow{10}{*}{0-shot}
        & MuSc & - & 80.4  & 97.6  & -  \\ \cmidrule(l){2-6}
        & Qwen2-VL  & 2B & - & - & 49.9\% \\ 
        & LLava-1.6 & 7B & - & - & 38.9\%  \\ 
        & LLaVA-OneVision-si & 8B & - & - & 60.8\%  \\ 
        & LLaVA-OneVision-ov   & 8B & - & - & 59.6\%  \\ 
        & Qwen2-VL  & 7B & - & - & 59.5\% \\ 
        & Qwen2-VL  & 72B & - & - & 59.6\%  \\ 
        & LLaVA-OneVision-ov  & 72B & - & - & 60.9\% \\ \cmidrule(l){2-6}
        & AnomalyGPT & 7B & 53.2 & 82.5  & 50.1\% \\
        & AnomalyGPT~(MuSc) & 7B & 80.4  & 97.6  & 52.3\% \\
        & {Myriad~(MuSc)} & 7B & 80.4  & 97.6  & \textbf{61.8\%} \\ \bottomrule
    \end{tabular}\vskip -0.1in

    \label{tab:pcbbank_fewshot}
\end{table}

\subsection{Qualitative Examples}
\label{sec:qualitative}
\begin{figure*}
\centering
\begin{subfigure}{.48\linewidth}
    \centering
    \includegraphics[width=0.95\linewidth]{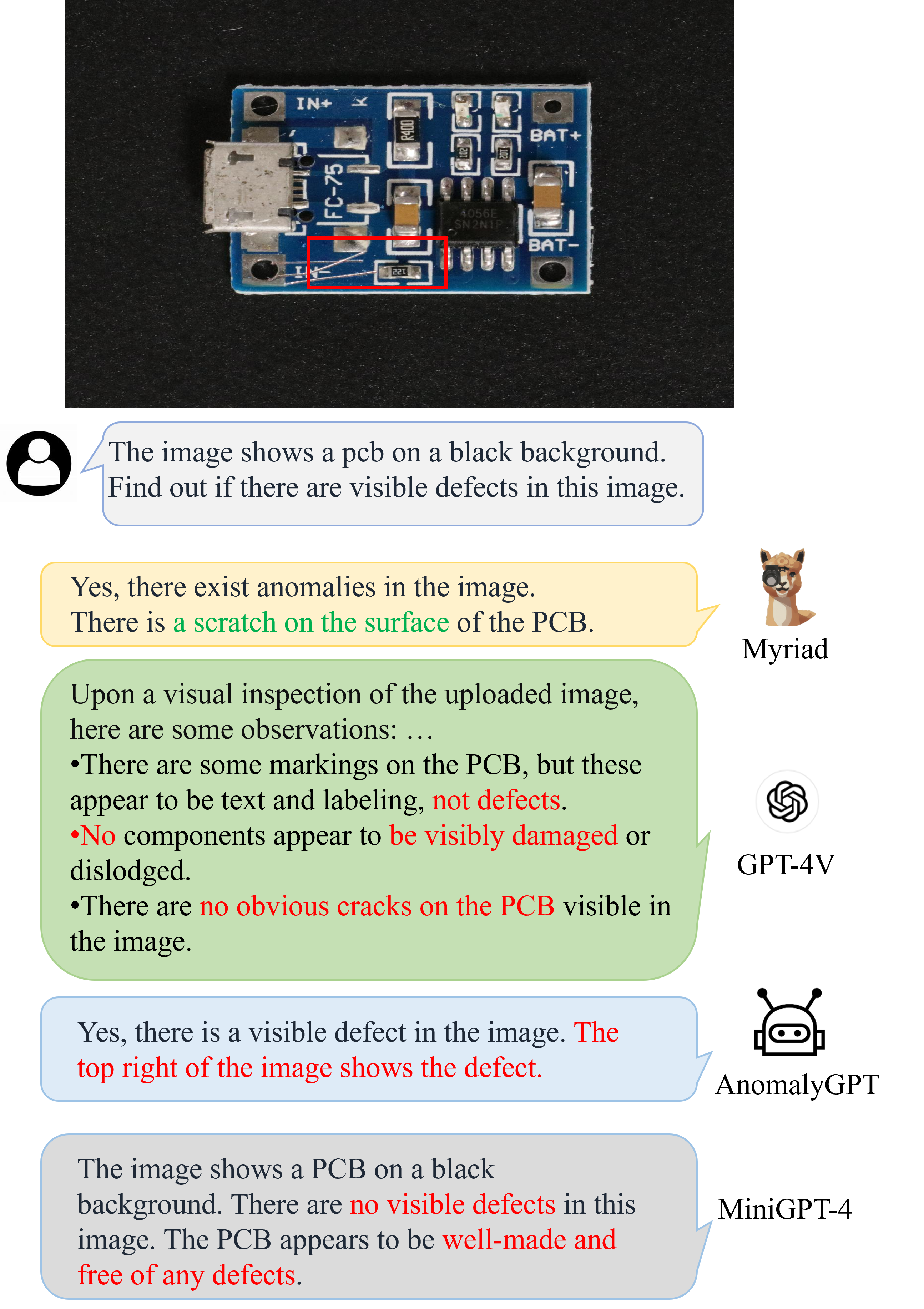}
    \caption{}
    \label{fig:qualitative-a}
\end{subfigure}
\hspace{.02\linewidth}
\begin{subfigure}{.48\linewidth}
    \centering
    \includegraphics[width=0.95\linewidth]{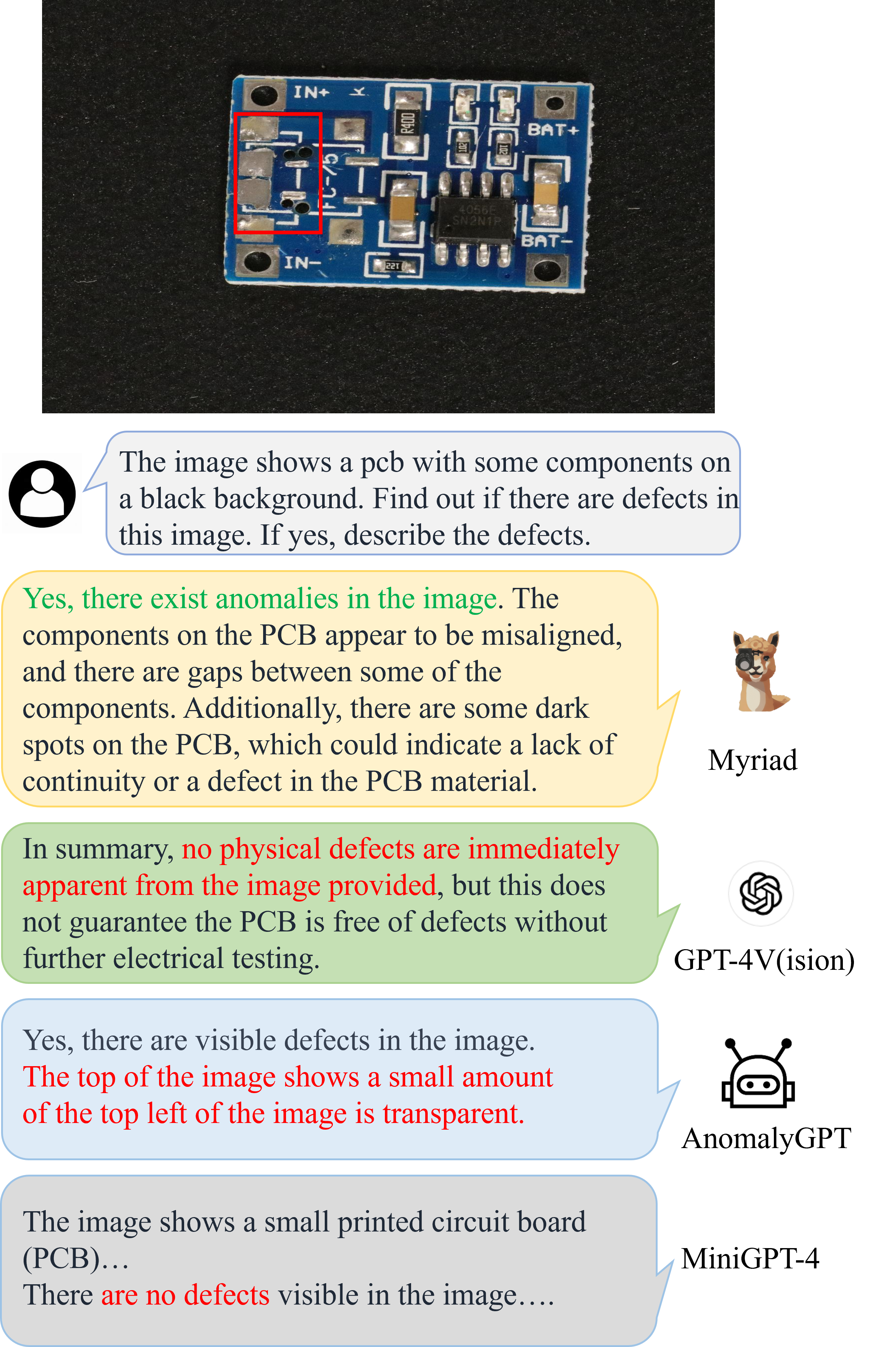}
    \caption{}
    \label{fig:qualitative-b}
\end{subfigure}
\begin{subfigure}{.48\linewidth}
    \centering
    \includegraphics[width=0.95\linewidth]{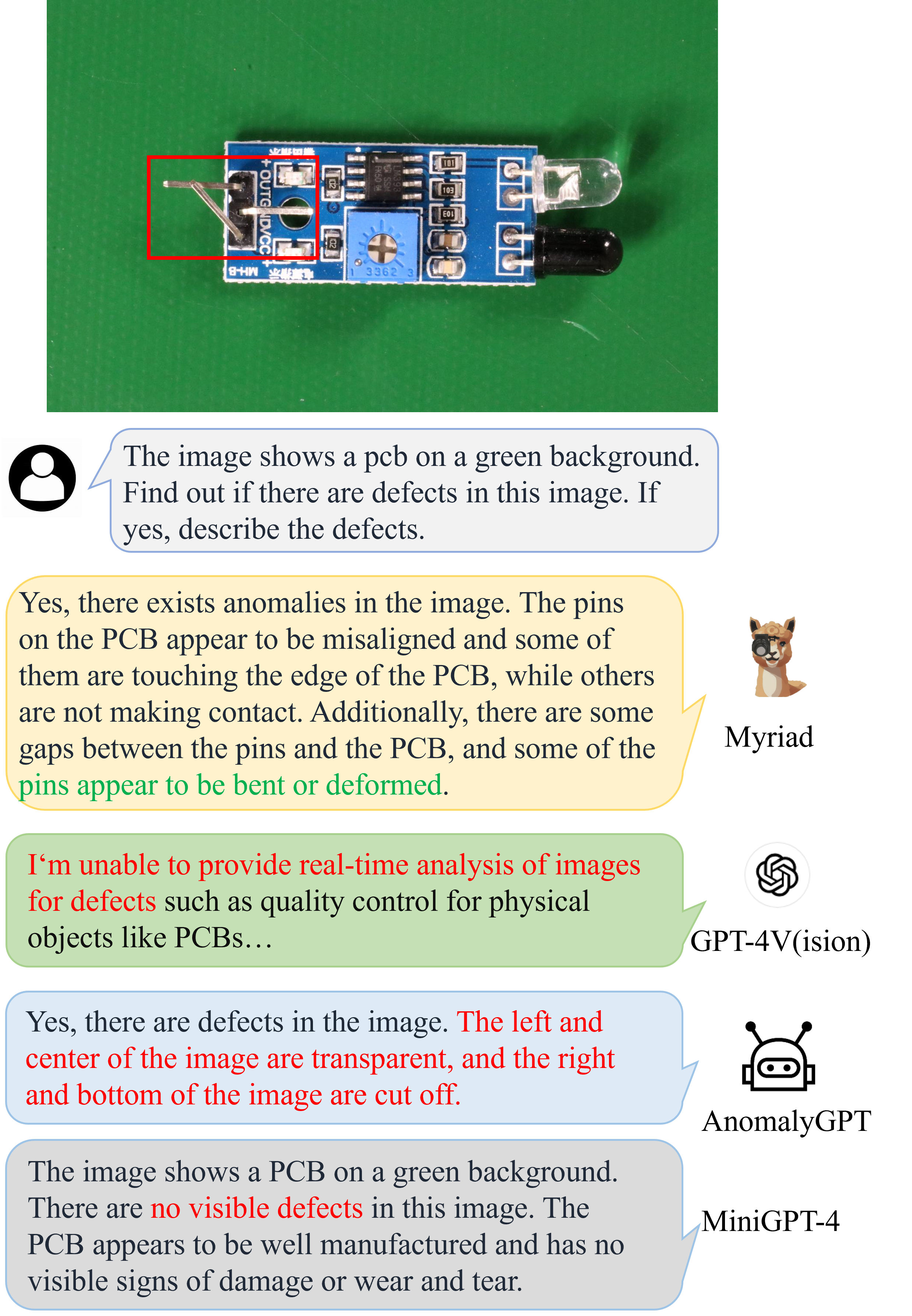}
    \caption{}
    \label{fig:qualitative-c}
\end{subfigure}
\hspace{.02\linewidth}
\begin{subfigure}{.48\linewidth}
    \centering
    \includegraphics[width=0.95\linewidth]{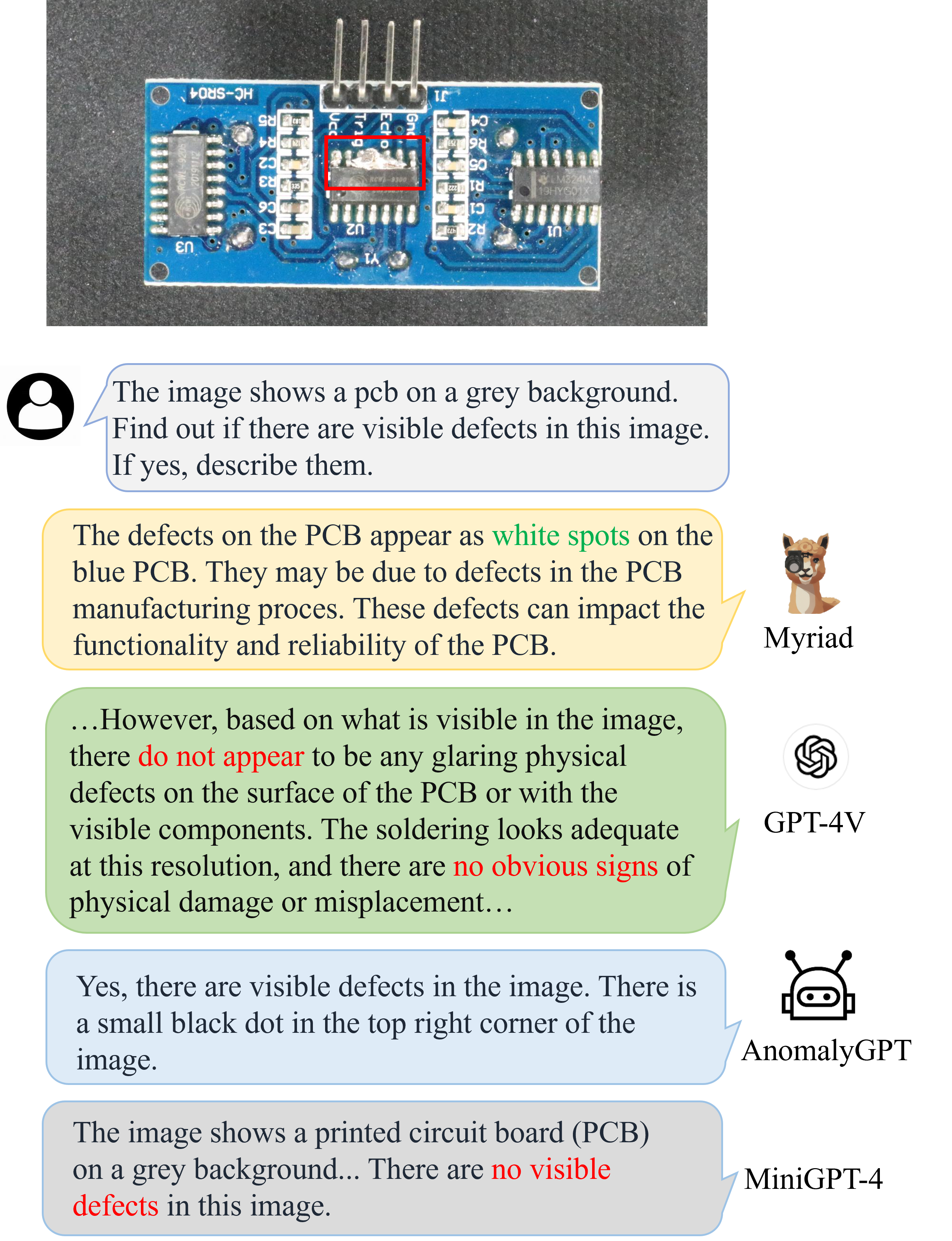}
    \caption{}
    \label{fig:qualitative-d}
\end{subfigure}
\caption{\textbf{The qualitative comparison between Myriad and the state-of-the-art LMMs.} Correct details are highlighted in \textbf{\color{green} green}, while incorrect details are marked in \textbf{\color{red} red}. The ground truth defects are highlighted with \textbf{\color{red} red} bounding boxes in the image. Best view in color version. }
\label{fig:qualitative}
\end{figure*}

We investigate the qualitative performance of Myriad in this section. 
First, we qualitatively compare Myriad with recent LMMs, including MiniGPT-4~\cite{minigpt4}, AnomalyGPT~\cite{adgpt}, and chatGPT-4V(ision)~\cite{gpt4v}. Specifically, we use the November 21, 2023 version for chatGPT-4V(ision). 
As shown in \cref{fig:qualitative}, GPT-4V provides more detailed image descriptions compared to MiniGPT-4. However, it still fails to detect existing anomalies in complex production environments.
While AnomalyGPT can identify the presence of anomalies, it often inaccurately describes them in detail.
In contrast, our method not only detects the existence of anomalies but also provides additional categorical information.

\begin{figure*}
\centering
\begin{subfigure}{.48\linewidth}
    \centering
    \includegraphics[width=0.95\linewidth]{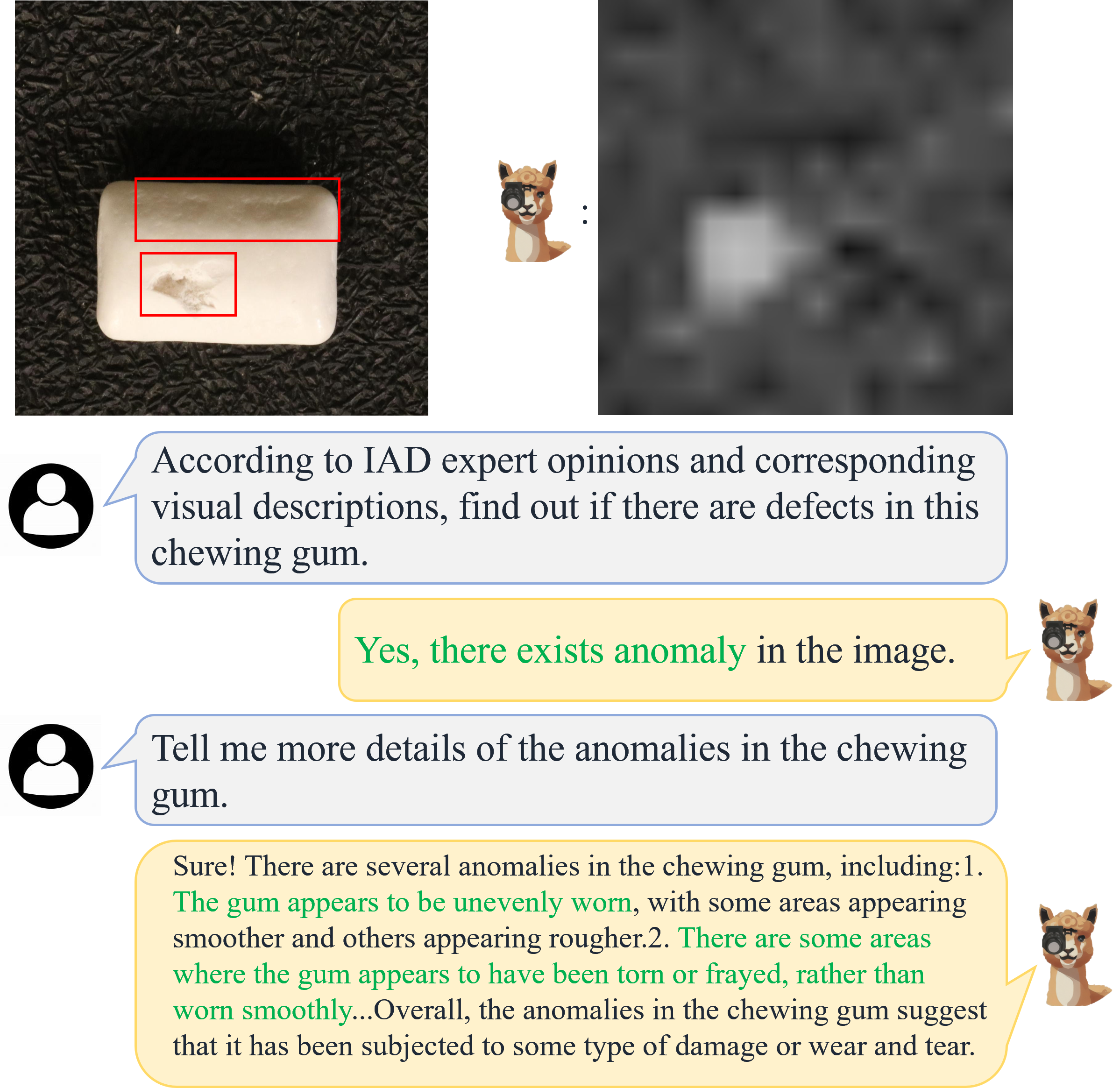}
    \caption{}
    \label{fig:flexible-a}
\end{subfigure}
\hspace{.02\linewidth}
\begin{subfigure}{.48\linewidth}
    \centering
    \includegraphics[width=0.95\linewidth]{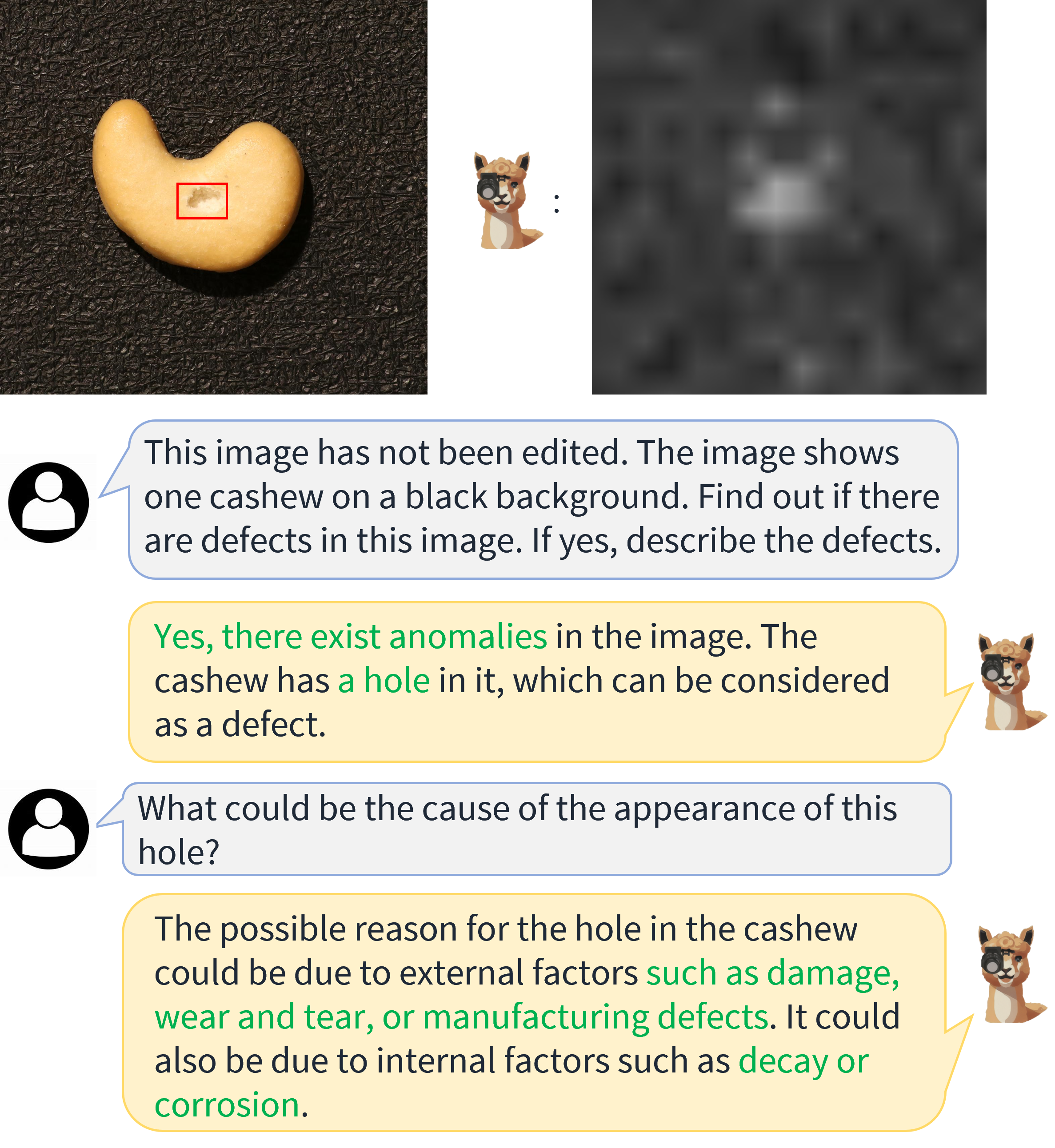}
    \caption{}
    \label{fig:flexible-b}
\end{subfigure}
\caption{\textbf{The illustration of instruction following ability.} The ground truth defects are highlighted with \textbf{\color{red} red} bounding boxes in the image. Best view in color version. }
\label{fig:flexible}
\end{figure*}

Compared to existing learning-based IAD approaches, Myriad demonstrates remarkable flexibility and instruction-following capabilities inherited from general domain LMMs. Specifically, Myriad can follow instructions to provide detailed descriptions of anomalies (see \cref{fig:flexible-a}). Moreover, users can leverage Myriad to identify the likely causes of these anomalies (see \cref{fig:flexible-b}).

Myriad further illustrates its flexibility by correcting previously inaccurate responses when provided with additional human expertise, such as information on product categories, process manuals, and common anomaly types (see \cref{fig:normal_hint}). 
The flexibility and instruction-following capabilities address the key limitations of current industrial anomaly detection methods, which incur high retraining costs when upgrading production lines or introducing new manufacturing technologies.
Consequently, constructing large multimodal models like Myriad holds significant research value for industrial anomaly detection.

\subsection{Ablation Studies}
\label{sec:ablation}

\vspace{0.5em}
\noindent\textbf{The effective of VE-Guided vision encoder.}
The VE-guided vision encoder plays a pivotal role in Myriad, as removing it causes a 2.2\% accuracy drop under the one-class setting and a 4.4\% drop under the one-shot setting. Specifically, using LoRRA or the visual prompt generator~(VPG) alone in the one-shot setting provides only marginal improvements of 0.2\% and 0.6\%, respectively. However, combining these two modules yields a performance gain of at least 3.8\% (80.0\% vs. 76.2\%), indicating that both expert guidance and IAD-specific visual features are crucial for detecting unseen product anomalies.

Under the one-class setting, improvements are largely smooth. LoRRA outperforms VPG, suggesting that LoRRA effectively learns product visual patterns for anomaly detection, while VPG can concentrate on the regions highlighted by external IAD models as potential defect areas.

\vspace{0.5em}
\noindent\textbf{The effective of textual prompt generator.}
The textual prompt generator offers comparable performance in both the one-class and one-shot settings. When using only expert guidance, AnomalyGPT outperforms Myriad (93.3\% vs. 92.0\% in the one-class setting and 77.4\% vs. 75.6\% in the one-shot setting), likely due to its ImageBind vision encoder, which has three times as many parameters as Myriad (1.2B vs. 0.4B). 
By integrating the VE-guided vision encoder, Myriad gains an additional 2.2\% and 4.4\% improvement in the one-class and one-shot settings, respectively, suggesting that expert prompts and vision modality information are complementary for industrial anomaly detection.

\vspace{0.5em}
\noindent\textbf{Comparison on computation cost and memory comsumption.}
We report the computational cost and memory efficiency of typical zero- and few-shot methods, including AprilGAN, MuSc, AnomalyGPT, and Myriad. In practice, Myriad can run on a single NVIDIA RTX 3090 which has 24GB memory. Moreover, Myriad can process up to 1.73 million samples per day, while AnomalyGPT handles only 860,000 images per day. As a result, Myriad is suitable for many industrial inspection scenarios that do not require strict real-time performance, such as final inspection stages.

\begin{table}[h]
    \centering
    \setlength{\tabcolsep}{1mm}
    \small
    \vspace{-1em}
    \captionsetup{font={footnotesize}}
    \caption{Comparison on computation and memory efficiency. Reported results are evaluated on one NVIDIA A800 80G.}
    \scalebox{0.7}{
    \begin{tabular}{@{}ccccc@{}}
        \hline
        Methods &   AprilGAN & MuSc & AnomalyGPT & Myriad\\ 
        \hline
        Training Time (h) &    0.3   &    -  &  $\sim$24  &   $\sim$24  \\
        Inference Time (s)   &   0.1      &   0.9   &  1.0 &  0.5    \\
        Inference Memory Consumption (GB) &  2.4      &   8.8   & 16.2  & 21.0      \\
        \hline
    \end{tabular}
    }
    \label{tab:cost}
    \vspace{-1em}
\end{table}

\begin{table}[]
    \centering
    \setlength{\tabcolsep}{1mm}
    \small
    \caption{Ablation studies on architecture.}
    \begin{tabular}{@{}ccccc@{}}
        \toprule
        \multirow{2}{*}[-0.5ex]{\makecell[c]{Textual Prompt Generator\\(TPG)}} &
        \multicolumn{2}{c}{VE-Guided Vision Encoder} &
        \multicolumn{2}{c}{Accuracy}
        \\ \cmidrule(lr){2-3} \cmidrule(lr){4-5}
        &  LoRRA &  VPG  & MVtec-AD (one-class) & VisA (1-shot) \\
        \midrule\\[-5mm]
        $\checkmark$ &              &              &   92.0   &  75.6     \\
        & $\checkmark$ & $\checkmark$ &   92.5   &  77.4    \\
        $\checkmark$ & $\checkmark$ &              &   93.5   &  75.8     \\
        $\checkmark$ &              & $\checkmark$ &   93.0   &  76.2     \\
        $\checkmark$ & $\checkmark$ & $\checkmark$ &\textbf{94.2} &\textbf{80.0} \\
        \bottomrule
    \end{tabular}
    \vskip -0.1in
    \label{tab:ablation}
\end{table}

\section{Conclusion}

In this paper, we introduce a novel large multimodal model, Myriad, designed to address industrial anomaly detection in a flexible and data-efficient manner. By leveraging existing industrial anomaly detection methods as ``vision experts,'' Myriad incorporates anomaly maps into large multimodal models to highlight critical regions, enhance the learning of anomalous features, and stimulate the underlying cognition of the experts. In this way, our approach benefits from both the specialized knowledge of IAD methods and the powerful generalization and instruction-following abilities of large multimodal models. Extensive experiments on MVTec-AD, VisA, and PCB Bank benchmarks demonstrate that Myriad achieves state-of-the-art performance under both one-class and few-shot settings. Moreover, thanks to the modularity of its architecture and the integration of vision experts, Myriad can be naturally extended to various real-world industrial scenarios without requiring substantial model redesign. We believe that this work opens promising avenues for bridging specialized expert models with large multimodal backbones, paving the way for more robust, adaptable, and efficient solutions in industrial anomaly detection.

\Acknowledgements{This work was supported in part by the National Key Research and Development Program of China under Grant No. 2022YFA1004100.}

{
    \small
    \bibliographystyle{IEEEtran}
    \bibliography{main}
}

\end{document}